\documentclass[10pt,twocolumn,letterpaper]{article}

\usepackage[pagenumbers]{cvpr} 

\usepackage{xcolor,pifont}
\usepackage{colortbl}
\usepackage{wrapfig}
\usepackage{overpic}
\usepackage{graphicx}
\usepackage{amsmath}
\usepackage{subcaption}
\usepackage{multirow}
\usepackage{tcolorbox}
\usepackage{booktabs}
\usepackage{cite}
\usepackage[font=small,labelfont=bf]{caption}
\newcommand{\best}[1]{\textbf{\color{red}{#1}}}
\newcommand{\second}[1]{\textcolor{blue}{#1}}
\usepackage{booktabs,multirow}
\usepackage[table]{xcolor}

\newcommand{\faceDetNotation}{$\mathcal{F}_{\text{face}}$\xspace}

\usepackage{xcolor}
\definecolor{citecolor}{RGB}{34,139,34}
\definecolor{lightred}{RGB}{255,100,100}
\definecolor{cell_bisque}{rgb}{1.0, 0.89, 0.77}
\definecolor{cell_blond}{rgb}{0.98, 0.94, 0.75}
\definecolor{cell_blue}{RGB}{155, 187, 228}
\definecolor{princetonorange}{rgb}{1.0, 0.56, 0.0}
\definecolor{pinkpearl}{rgb}{0.91, 0.67, 0.81}
\definecolor{mossgreen}{rgb}{0.68, 0.87, 0.68}
\definecolor{cadmiumgreen}{rgb}{0.0, 0.42, 0.24}
\definecolor{brightmaroon}{rgb}{0.76, 0.13, 0.28}
\definecolor{calpolypomonagreen}{rgb}{0.12, 0.35, 0.17}
\definecolor{darkseagreen}{rgb}{0.56, 0.74, 0.56}
\definecolor{armygreen}{rgb}{0.29, 0.33, 0.13}
\definecolor{azure}{rgb}{0.0, 0.5, 1.0}
\definecolor{denim}{rgb}{0.08, 0.38, 0.84}
\definecolor{bananayellow}{rgb}{1.0, 0.88, 0.21}

\newcommand{\Paragraph}[1]{\vspace{-0mm}\noindent\textbf{#1.}\hspace{0mm}}

\newcommand{\SubSection}[1]{\vspace{-0mm} \subsection{#1} \vspace{-0mm}}

\definecolor{cvprblue}{rgb}{0.21,0.49,0.74}
\definecolor{linkcolor}{rgb}{0.93, 0.11, 0.14}
\definecolor{citecolor}{rgb}{0, 113, 188}
\usepackage[pagebackref,breaklinks,colorlinks,linkcolor=linkcolor,citecolor=cvprblue]{hyperref}
            
\def\paperID{} 
\def\confName{CVPR}
\def\confYear{2026}

\title{On the Holistic Approach for Detecting Human Image Forgery}

\begin{document}

\author{Xiao Guo, Jie Zhu, Anil Jain, Xiaoming Liu\\
Michigan State University\\
{\tt\small \{guoxia11, zhujie4, jain, liuxm\}@msu.edu}
}

\maketitle
\begin{abstract}
The rapid advancement of AI-generated content (AIGC) has escalated the threat of deepfakes, from facial manipulations to the synthesis of entire photorealistic human bodies.
However, existing detection methods remain fragmented, specializing either in facial-region forgeries or full-body synthetic images, and consequently fail to generalize across the full spectrum of human image manipulations.
We introduce HuForDet, a holistic framework for human image forgery detection, which features a dual-branch architecture comprising: (1) a face forgery detection branch that employs heterogeneous experts operating in both RGB and frequency domains, including an adaptive Laplacian-of-Gaussian (LoG) module designed to capture artifacts ranging from fine-grained blending boundaries to coarse-scale texture irregularities; and (2) a contextualized forgery detection branch that leverages a Multi-Modal Large Language Model (MLLM) to analyze full-body semantic consistency, enhanced with a confidence estimation mechanism that dynamically weights its contribution during feature fusion.
We curate a human image forgery (HuFor) dataset that unifies existing face forgery data with a new corpus of full-body synthetic humans. 
Extensive experiments show that our HuForDet achieves state-of-the-art forgery detection performance and superior robustness across diverse human image forgeries. 
\end{abstract}
\begin{figure*}[t]
  \centering
    \begin{overpic}[width=1\linewidth]{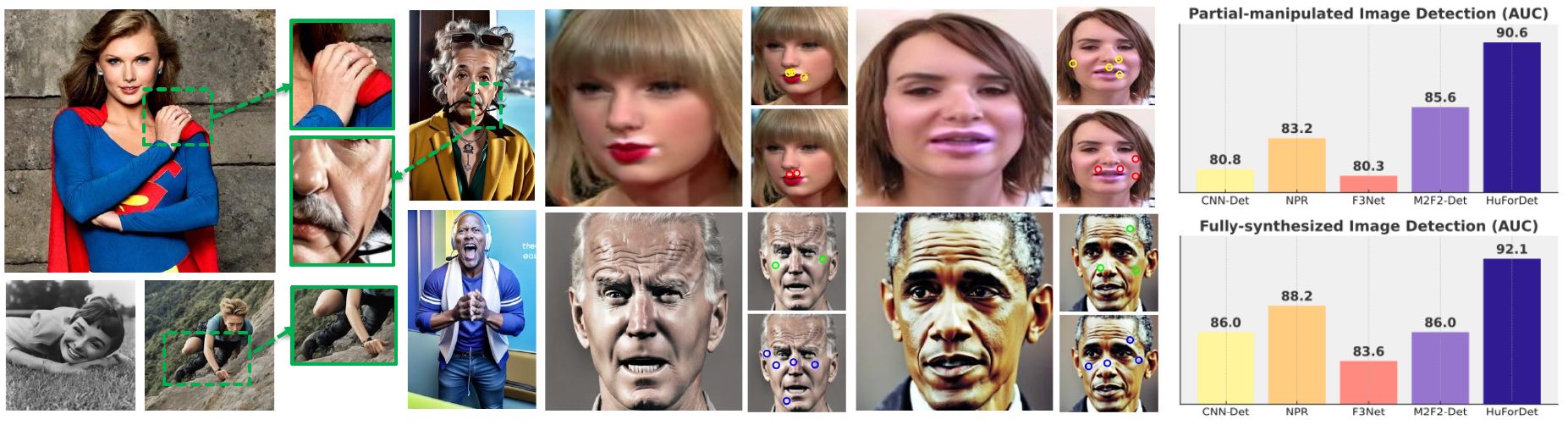}
    \put(19,0){(a)}
    \put(53.3,0){(b)}
    \put(86.3,0){(c)}
    \end{overpic}
    \vspace{-6mm}
    \caption{\textbf{(a)} Beyond facial forgeries, AIGC methods enable the synthesis of full-body human images, introducing distinctive anatomical anomalies such as an additional finger, unnaturally smooth skin, and three-legged artifacts.
    \textbf{(b)} The Laplacian of Gaussian (LoG) operator is an effective blob detector for identifying regions with rapid intensity changes, which often correspond to facial forgery artifacts. 
    However, conventional LoG-based detectors~\cite{hifi_net_xiaoguo,masi2020two} rely on a fixed scale parameter $\sigma$, capturing only a narrow subset of these artifact patterns. 
    Colored overlays show LoG blob detections at different scales $\sigma$: \textcolor{bananayellow}{\textbf{yellow}} ($\sigma{=}1$) and \textcolor{red}{red} ($\sigma{=}5$) in the first row highlight unnaturally bright mouth regions and blending artifacts; \textcolor{green}{green} ($\sigma{=}9$) and \textcolor{blue}{blue} ($\sigma{=}13$) in the second row emphasize abnormal skin textures.
    Our adaptive LoG (Sec.~\ref{subsec:gated_svle}) overcomes this limitation by learning optimal scales, adaptive to different spatial locations.
    \textbf{(c)} Our proposed \textbf{HuForDet} (Fig.~\ref{fig_archi}) achieves state-of-the-art performance on detecting both partial-manipulation (\textit{e.g.}, face-swap) and fully synthesized forgeries (\textit{e.g.}, GAN-generated faces, diffusion-generated full-body images) on our proposed HuFor dataset (Sec.~\ref{sec:dataset}). \vspace{-5mm}}
    \label{fig_overall_2}
\end{figure*}\vspace{-5mm}
\section{Introduction}
The proliferation of deepfakes poses a serious threat to the trustworthiness of human identity, making media forensics a critical research topic.
Traditional forgery techniques primarily focused on image editing or face-swapping algorithms to manipulate a subject’s identity, often altering facial regions while preserving the overall image context. 
However, recent advances in AI-generated content (AIGC)~\cite{goodfellow2014generative, choi2018stargan, karras2019style,rombach2022high,guo2024dense} enable the synthesis of photorealistic, full-body human images, where the forgeries span the entire human figure, as illustrated in Fig.~\ref{fig_overall_2}\textcolor{red}{a}.
Consequently, we need to address a new problem: detecting human image forgeries, 
regardless of whether manipulations or syntheses occur on the face or other body parts. 

However, existing detection methods are poorly suited for human image forgery detection. 
First, conventional deepfake detection methods~\cite{chen2022self,li2020face,liexposing,shiohara2022detecting,qian2020thinking,gu2022exploiting} operate on cropped faces, rendering them ineffective when applied to full-body synthetic images that have forgery outside face regions.  
On the other hand, AIGC image detectors~\cite{cozzolino2024raising,ojha2023towards,sha2023fake} process the entire image and can be unreliable when the manipulated face region constitutes only a small part of the overall image.
Therefore, we propose \textit{HuForDet}, a holistic dual-branch architecture for human image forgery detection. 
One branch specializes in face-region analysis, leveraging a mixture-of-experts (MoE) design that fuses RGB and frequency-domain features to capture diverse facial forgeries. 
The second branch performs contextualized full-body analysis, utilizing semantic cues to identify human anatomical distortions (\textit{e.g.}, broken fingers, unnatural body shapes). 
By fusing representations from both branches, HuForDet effectively detects a wide spectrum of human image forgeries.

The face forgery detection branch comprises experts specializing in RGB and frequency domains, motivated by an observation that different generation processes leave distinct forgery traces. 
\underline{\textit{First}}, partial manipulation techniques, \textit{e.g.}, face-swapping, introduce blending artifacts around manipulation boundaries~\cite{qian2020thinking,gu2022exploiting,liu2021spatial,luo2021generalizing,wang2023dynamic}.
To capture these patterns, we employ the Laplacian of Gaussian (LoG) operator~\cite{burt1987laplacian} to amplify high-frequency forgery cues in forged faces.
However, existing frequency-based methods~\cite{li2004live,liu2021spatial,luo2021generalizing,wang2023dynamic}, including those using LoG, rely on fixed filters. 
This can make them inherently limited when confronting forgeries where artifacts exhibit significant multi-scale and spatial variations, as visualized in Fig.~\ref{fig_overall_2}\textcolor{red}{b}.
We therefore introduce an adaptive LoG block (adaLoG) --- a learnable, multi-scale frequency-domain expert that dynamically captures frequency features across varying scales. 
To ensure comprehensive coverage, we deploy two adaLoG blocks as complementary experts, specializing in fine-grained blending boundaries to coarse-scale texture irregularities.
\underline{\textit{Secondly}}, fully-synthesized faces from GANs or diffusion models exhibit fewer blending artifacts but often contain structural abnormalities like implausible facial geometry and misalignments~\cite{chen2022self,li2020face,liexposing,shiohara2022detecting,zhao2021learning,xiao_hifinet_plusplus}. 
These patterns are more effectively captured by RGB domain experts, which learn spatial relationships and dependencies between facial components.
Therefore, by incorporating both frequency and RGB domain experts, the face forgery detection branch ensures comprehensive coverage of diverse forgery patterns in facial regions.

\begin{figure*}[t]
  \centering
    \includegraphics[width=0.9\linewidth]{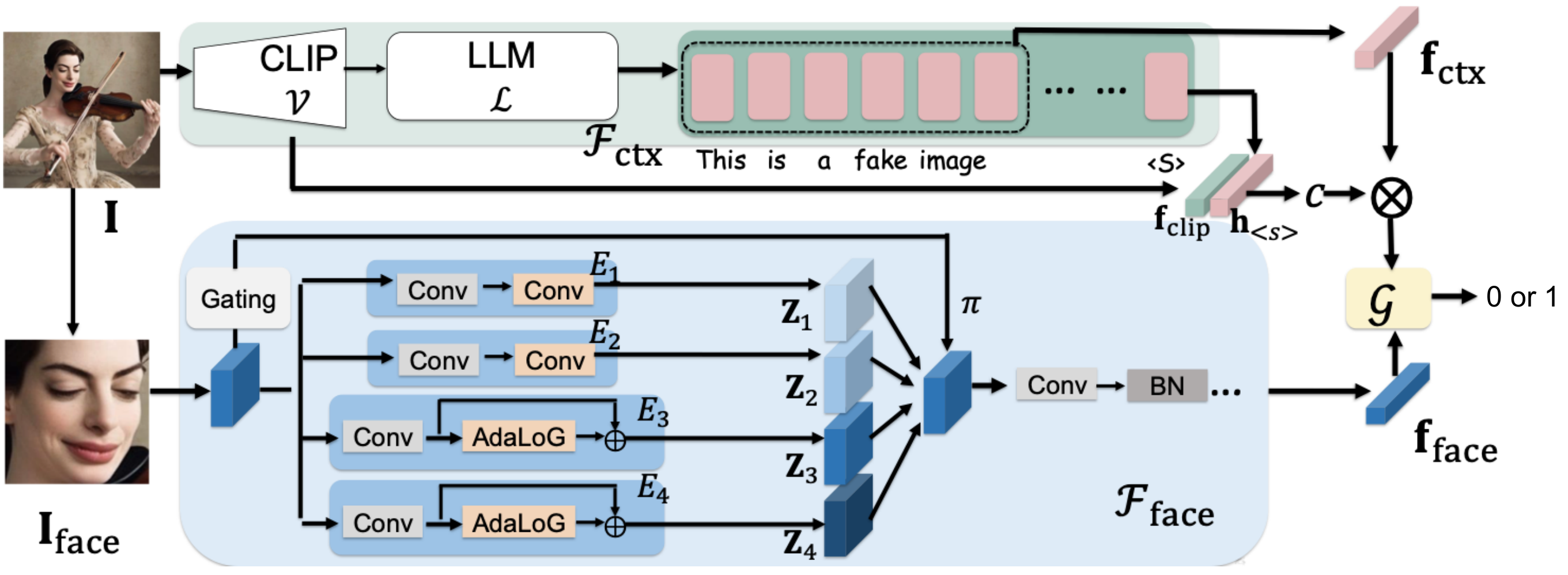}
    \caption{
    Our HuForDet comprises two branches: a \emph{face forgery detection branch} (\faceDetNotation) and a \emph{contextualized forgery detection branch} ($\mathcal{F}_{\text{ctx}}$), which are introduced in Sec.~\ref{sec:face_branch} and Sec.~\ref{subsec:context_branch}, respectively.
    Specifically, \faceDetNotation analyzes cropped face regions $\mathbf{I}_{\text{face}}$ using heterogeneous RGB spatial (\textit{i.e.}, $E_1$ and $E_2$) and frequency domain (\textit{i.e.}, $E_3$ and $E_4$) experts, and then it generates a facial forgery representation $\mathbf f_{\text{face}}\in\mathbb R^{d}$. 
    Also, $\mathcal{F}_{\text{ctx}}$ processes the input image $\mathbf{I}$ to produce a contexualized forgery representation $\mathbf f_{\text{ctx}}\in\mathbb R^{d}$ and a self-assessed confidence $c\in[0,1]$. 
    A confidence-aware fusion module $\mathcal G$ aggregates $\mathbf f_{\text{face}}$ and $\mathbf f_{\text{ctx}}$ to produce a holistic representation for the final forgery prediction.
    \vspace{-5mm}}
    \label{fig_archi}
\end{figure*}

HuForDet also uses a contextualized forgery detection branch to analyze the full-body image for global semantic forgery clues (\textit{e.g.}, implausible limb articulations, unnatural human skins), which provides complementary detection power to the face forgery detection branch.
While this branch leverages MLLMs' comprehension capabilities, it inherits a fundamental limitation: the tendency to generate erroneous outputs when visual artifacts are subtle.
To mitigate this, we train the contextualized forgery detection branch to conclude its output with a special token. 
The hidden state of this token provides a compact representation of the model's self-assessed certainty based on its reasoning. 
Also, we condition it on the global image context from the vision encoder, and then regress it to a confidence score, as depicted in Fig.~\ref{fig_archi}.
This score informs the fusion mechanism how much learned contextualized forgery features contribute, depending on input forgery categories.

To facilitate human image forgery detection research, we construct HuFor, a large-scale dataset for human image forgery detection, detailed in Sec.~\ref{sec:dataset}.
HuFor combines FaceForensics++~\cite{rossler2019faceforensics++v3} and UniAttack+~\cite{liu2025benchmarking} datasets, which cover $28$ face forgery types, including both partial manipulation and full synthesis. 
Furthermore, we expand the HuFor dataset by using state-of-the-art (SoTA) personalized diffusion models~\cite{wang2024instantid,li2024photomaker,ye2023ip}, which generate a novel corpus of high-resolution, full-body human forgeries in diverse contexts. 
HuFor provides the necessary foundation for training and evaluating models on the full spectrum of human image forgeries.
Empirically, our HuForDet achieves SoTA performance on the HuFor dataset and exhibits remarkable generalization capabilities across different forgery types (Fig.~\ref{fig_overall_2}\textcolor{red}{c}). 
In summary, our contributions are:

$\diamond$ We propose HuForDet, a holistic human image forgery detection method that detects forgeries through joint analysis of local facial manipulation traces and anomalous human body constructions in full-body images.

$\diamond$ We design a face forgery detection branch with heterogeneous experts, containing a novel adaptive LoG block for a comprehensive frequency-domain representation, enabling robust detection of diverse forgery patterns.

$\diamond$ HuForDet contains a contextualized forgery detection branch, which not only identifies semantic human image generation artifacts but also outputs a confidence score to guide the fusion of its output into the final representation. 

$\diamond$  The proposed HuForDet achieves SoTA performance on the human image forgery (HuFor) dataset, which integrates existing face forgery datasets with a newly curated set of full-body synthetic human images.

\section{Related Works}
\Paragraph{Human Image Forgery Detection}
Human image forgeries manifest as either partial manipulations (e.g., face-swapping) or full syntheses (e.g., GAN-generated faces, diffusion-generated full-body images).
Prior detection methods mainly target these challenges in isolation: 
Most conventional forgery detectors~\cite{li2020face,liexposing,shiohara2022detecting,zhao2021learning,wang2021representative,tan2024data} focus on identifying partial facial manipulations while largely overlooking full-body synthesis.
Conversely, recent AIGC detection methods detect fully synthesized images. 
These include techniques that leverage frequency domain analysis~\cite{wang2020cnn,frank2020leveraging}, reconstruction errors of diffusion models~\cite{wang2023dire}, pre-trained vision-language models for generalization~\cite{ojha2023towards}, and local artifact analysis~\cite{zhong2023patchcraft,tan2024rethinking}. 
Methods like HumanSAM~\cite{liu2025humansam} and AvatarShield~\cite{xu2025avatarshield} address human forgeries but focus exclusively on fully-synthesized video content. 
These methods fail to address partially forged images where manipulations occur only in localized regions. 
To bridge this gap, we propose HuForDet, a holistic forgery detection method for localized facial manipulations to full-body synthetic artifacts.

\Paragraph{Mixture of Experts}
The idea of mixture of experts (MoE) represents an effective machine learning paradigm~\cite{jacobs1991adaptive,jordan1994hierarchical} for tackling complex tasks through the ensemble of specialized experts, each of which focuses on a specific subspace of the input distribution, thereby efficiently modeling heterogeneous data patterns. 
Recently, MoE frameworks have been used in diverse fields such as natural language processing~\cite{fedus2022switch,jiang2024mixtral,komatsuzaki2022sparse,lepikhin2020gshard,shazeer2017outrageously}, computer vision~\cite{abbas2020biased,ahmed2016network,chen2024eve,chen2023adamv,eigen2013learning,ge2020self}, and biometrics~\cite{dai2021generalizable,jawade2024proxyfusion,wang2021multi,wang2025decoupled,su2025hamobe,zhu2025quality,wen2015face}.
One similar work, MoE-FFD~\cite{kong2025moe}, introduces a MoE module with homogeneous experts within a ViT-based architecture for forged faces, while we use a set of heterogeneous experts tailored for distinct traces, targeting more diverse forgery types. 
Also, MoE-FFD repeatedly applies MoE across layers, but our work only has MoE in the early layers of the model, yielding better computational efficiency.

\Paragraph{Multimodal Large Language Models}
MLLMs~\cite{yin2023survey,li2022blip,li2023blip,liu2024visual} use generative capabilities of LLMs to obtain impressive performance across a wide range of tasks~\cite{touvron2023llama,zhang2022opt,zhang2024vision,kamali2024nesycoco}. 
For example, early studies  generate text-based content grounded on image~\cite{zhang2023vln,zhang2024navhint}, video, and audio~\cite{liu2024visual,li2023videochat,awadalla2023openflamingo,zhang2023video,deshmukh2023pengi,wu2023cheap}. 
Recently, MLLM-based methods have been adopted in the deepfake detection community~\cite{zhang2024common,guo2025rethinking,xiufeng_lamma_detection,peng2025mllm,zhou2025aigi,tan2025veritas}, identifying appearances that do not obey the common sense or laws of physics.
However, these works do not address MLLMs' hallucination issues and other inherent limitations in identifying subtle forgery traces. 
To address this, our contextualized forgery detection branch outputs a confidence score that guides its output to fuse into final forgery representations, depending on input forgery attributes.
\section{Method}
\label{sec:method_overall}

\subsection{Overview}
\label{sec:preliminary}

Let us denote an input image as $\mathbf{I} \in \mathbb{R}^{H \times W \times 3}$. Our HuForDet learns a mapping function $\mathcal{F}$ that predicts a forgery probability $y \in [0, 1]$, where $y=1$ indicates a forgery. 
Our HuForDet (Fig.~\ref{fig_archi}) consists of two major components:
(a) \textbf{The Face Forgery Detection Branch} ($\mathcal{F}_{\text{face}}$) takes as input a cropped facial region $\mathbf{I}_{\text{face}} = \mathcal{C}(\mathbf{I})$,  with $\mathcal{C}$ being the face cropping function.
It processes this region to extract a discriminative feature representation $\mathbf{f}_{\text{face}} \in \mathbb{R}^{d}$ focused on local face forgery traces.
(b) \textbf{The Contextualized Forgery Detection Branch} ($\mathcal{F}_{\text{ctx}}$) takes the entire image $\mathbf{I}$ as input. It has two outputs (i) a semantic feature embedding $\mathbf{f}_{\text{ctx}} \in \mathbb{R}^{d}$ that encodes high-level, global cues of forgeries, and (ii) a self-assessed confidence score $c \in [0, 1]$ that quantifies the certainty of its own assessment.
Then, we use a confidence-aware fusion network $\mathcal{G}$ that integrates the complementary information from both branches:
\begin{equation}
y = \mathcal{G}( \mathbf{f}_{\text{face}}, \mathbf{f}_{\text{ctx}}, c; \Theta_{\mathcal{G}} ), \quad \text{where} \quad 
\begin{cases} 
\mathbf{f}_{\text{face}} = \mathcal{F}_{\text{face}}(\mathcal{C}(\mathbf{I})), \\
(\mathbf{f}_{\text{ctx}}, c) = \mathcal{F}_{\text{ctx}}(\mathbf{I}).
\end{cases}
\label{eq:fusion}
\end{equation}
where $\Theta_{\mathcal{G}}$ represents parameters of the fusion network.

\SubSection{Face Forgery Detection Branch}
\label{sec:face_branch}

As shown in Fig.~\ref{fig_archi}, $\mathcal{F}_{\text{face}}$ uses four heterogeneous experts to analyze the input face region $\mathbf{I}_{\text{face}}$. 
Formally, let us denote input feature map as $\mathbf{X} \in \mathbb{R}^{C \times H \times W}$
and four heterogeneous experts as $\{E_1, E_2, E_3, E_4\}$.
Two spatial RGB domain experts (\textit{i.e.}, $E_1$ and $E_2$) use standard convolutional blocks with different kernel sizes (\textit{i.e.}, 3$\times$3 and 9$\times$9), which help capture structural irregularities from fine-grained and broader contexts, respectively. 
The frequency-domain experts $E_3$ and $E_4$ are based on adaptive Laplacian of Gaussian (adaLoG) blocks, detailed in Sec.~\ref{subsec:gated_svle}. 
Thus, $E_3$ operates on a finer scale ($\sigma \in {1,4,7}$) to highlight sharp, high-frequency cues like blending boundaries. In comparison, $E_4$ operates on a coarser scale ($\sigma \in {9,12,15}$) to detect anomalies such as unnatural smoothness. 
Both experts work collaboratively to analyze the frequency domain across different bandwidths. 
Formally, let $\mathbf{Z}_k$ denote the output of expert $E_k$, where $\mathbf{Z}_1, \mathbf{Z}_2$ are from spatial experts and $\mathbf{Z}_3, \mathbf{Z}_4$ are from frequency-domain experts. 
Subsequently, a gating network $G$, implemented as a $1\times1$ convolution followed by global average pooling and softmax, computes gate scores $\boldsymbol{\pi} \in \mathbb{R}^4$ that weight four experts' outputs:
\begin{align}
\boldsymbol{\pi} &= \mathrm{Softmax}(G(\mathbf{X}; \Theta_G)),
\mathbf{X}_{\text{moe}} &= \sum_{k=1}^4 \pi_k \cdot \mathbf{Z}_k.
\label{eq:moe}
\end{align}
The resulting feature map $\mathbf{X}_{\text{moe}}$ is passed through remaining layers of $\mathcal{F}_{\text{face}}$, which outputs the final feature vector $\mathbf{f}_{\text{face}}$.

\SubSection{Adaptive LoG Block}
\label{subsec:gated_svle}
\paragraph{LoG Operator Approximation} 
Given an input feature map $\mathbf{X} \in \mathbb{R}^{C \times H \times W}$, an adaptive LoG block generates a multi-scale representation via $K$ Gaussian smoothing operations $G_{\sigma_k}(\cdot)$ with distinct scales $\sigma_k$, producing filtered outputs:
\begin{equation}
\mathbf{Y}_k = \mathbf{X} - G_{\sigma_k}(\mathbf{X}) \quad \text{for } k=1, \dots, K.
\end{equation}
This operation provides a mathematical approximation of the LoG operator.
More formally, as established in scale-space theory~\cite{lindeberg2013scale,witkin1983scale}, the LoG operator is proportional to the scale derivative of the Gaussian-filtered image:
\begin{equation}
\text{LoG}(\mathbf{X}) = \frac{1}{2}\nabla^2 (G_{\sigma} * \mathbf{X}) \propto \frac{\partial}{\partial \sigma} (G_{\sigma} * \mathbf{X}).
\end{equation}

We approximate this continuous derivative using a finite-difference scheme that evaluates the change from scale $\sigma=0$ (the identity operation, yielding $\mathbf{X}$) to scale $\sigma_k$:
\begin{equation}
\frac{\partial}{\partial \sigma} (G_{\sigma} * \mathbf{X}) \approx \frac{G_{\sigma_k} * \mathbf{X} - G_0 * \mathbf{X}}{\sigma_k - 0} = \frac{G_{\sigma_k} * \mathbf{X} - \mathbf{X}}{\sigma_k}.
\end{equation}
Rearranging this approximation reveals the precise relationship to our operation:
\begin{align}
\mathbf{X} - G_{\sigma_k} * \mathbf{X}
&\approx -\,\sigma_k\,\frac{\partial}{\partial \sigma}\!\big(G_{\sigma} * \mathbf{X}\big) \\
&\approx -\,\sigma_k\,\mathrm{LoG}(\mathbf{X}).
\end{align}

This demonstrates that $\mathbf{X} - G_{\sigma_k}(\mathbf{X})$ provides a scaled approximation of the negative LoG response. 
This approximation preserves the essential blob-detection characteristics of the LoG operator while remaining computationally efficient and fully differentiable, making it ideal for integration into our end-to-end learnable architecture. 

\paragraph{Multi-Scale Adaptive Fusion} 
The fusion of these multi-scale representations is achieved by a controller network $g_\phi$, which enables content-aware adaptation.
Formally, $g_\phi$ analyzes input content and outputs a comprehensive decision map: $\mathbf{O}_{\text{raw}} = g_\phi(\mathbf{X}) \in \mathbb{R}^{(K+1) \times H \times W}$. 
We derive two types of control signals from $\mathbf{O}_{\text{raw}}$: blend weights $\mathbf{c}_k \in \mathbb{R}^{1 \times H \times W}$ and a gating map $\boldsymbol{\lambda} \in \mathbb{R}^{1 \times H \times W}$.
Specifically, $\mathbf{c}_k$ for each filter $k$ are obtained by applying Softmax across the first $K$ channels of $\mathbf{O}_{\text{raw}}$:
\begin{equation}
\mathbf{c}_k(h, w) = \frac{\exp(\mathbf{O}_{\text{raw}}[k, h, w])}{\sum_{j=1}^{K} \exp(\mathbf{O}_{\text{raw}}[j, h, w])}.
\end{equation}
Simultaneously, $\boldsymbol{\lambda}$ is derived by applying \texttt{Sigmoid} to $\mathbf{O}_{\text{raw}}$'s $(K+1)$-th channel:
\begin{equation}
\boldsymbol{\lambda}(h, w) = \frac{1}{1 + \exp(-\mathbf{O}_{\text{raw}}[K+1, h, w])}.
\end{equation}
Then, a composite feature map $\mathbf{Y}_{\text{comp}} = \sum_{k=1}^{K} \mathbf{c}_k \odot \mathbf{Y}_k$ is merged with the original input via the gating mechanism: $\mathbf{Z} = (1 - \boldsymbol{\lambda}) \odot \mathbf{X} + \boldsymbol{\lambda} \odot \mathbf{Y}_{\text{comp}}$,
where $\odot$ denotes element-wise multiplication and $\mathbf{Z}$ represents the final output of the adaptive LoG block, serving as either $\mathbf{Z}_3$ or $\mathbf{Z}_4$ in the mixture of experts framework.
\subsection{Contextualized Forgery Detection Branch}
\label{subsec:context_branch}

The HuForDet leverages a contextualized forgery detection branch, \textit{i.e.}, $\mathcal{F}_{\text{ctx}}$, to identify high-level semantic inconsistencies, which consists of a vision encoder $\mathcal{V}$ and a large language model (LLM) $\mathcal{L}$. 
The vision encoder first converts the image into a sequence of visual tokens: $\mathbf{T}_{\text{visual}} = \mathcal{V}(\mathbf{I})$.
These tokens are then combined with a system prompt $\mathbf{P}_{\text{sys}}$ and user query $\mathbf{P}_{\text{user}}$ to form the input sequence for $\mathcal{L}$:
$\mathbf{T}_{\text{input}} = [\mathrm{Tokenize}(\mathbf{P}_{\text{sys}}), \mathbf{T}_{\text{visual}}, \mathrm{Tokenize}(\mathbf{P}_{\text{user}})]$.
The $\mathcal{F}_{\text{ctx}}$ uses this sequence to generate two critical components: contextualized forgery representations and a confidence score, denoted as $\mathbf{f}_{\text{ctx}}$ and $c$, respectively.
As depicted in Fig.~\ref{fig_archi}, $\mathcal{F}_{\text{ctx}}$ uses $\mathcal{L}$ to autoregressively generate a sequence of text tokens.
We obtain these tokens' corresponding embeddings $\mathbf{W} = [\mathbf{w}_1, \mathbf{w}_2, ..., \mathbf{w}_N]$ and then aggregate $\mathbf{W}$ (\textit{e.g.}, via max pooling) into a holistic text representation $\overline{\mathbf{W}}$, which encapsulates the semantic rationale for the forgery decision. 
This representation is then projected into $\mathbf{f}_{\text{ctx}}$ via
a \texttt{MLP},
namely $\mathbf{f}_{\text{ctx}} = \texttt{MLP}(\overline{\mathbf{W}})$.

The $\mathcal{L}$ is trained to conclude its textual response with the special \texttt{<s>} token. 
The final hidden state of this token from $\mathcal{L}$'s last layer, denoted as  $\mathbf{h}_{<s>} \in \mathbb{R}^{d}$, serves as a compact representation of the model's self-assessed certainty based on its reasoning. 
To further ground this confidence estimation in the visual input, we condition it on the global image context, \textit{i.e.}, $\mathbf{f}_{\text{clip}}$, obtained from the CLIP vision encoder.
The joint representation $[\mathbf{h}{<s>}; \mathbf{f}_{\text{clip}}]$ is then regressed into a scalar confidence score through an MLP with \texttt{Sigmoid}:
\begin{equation}
c = \texttt{Sigmoid} \left( \texttt{MLP}([\mathbf{h}_{<s>}; \mathbf{f}_{\text{clip}}]) \right).
\label{eq:confidence_score}
\end{equation}
This score $c$ dynamically governs the contribution of $\mathcal{F}_{\text{ctx}}$ in the final fusion network (Eq. \ref{eq:fusion}), reducing its influence when its prediction is uncertain and thus robustly mitigating the risk of relying on hallucinated rationales.
\begin{figure}[t]
    \centering
     \includegraphics[width=1\linewidth]{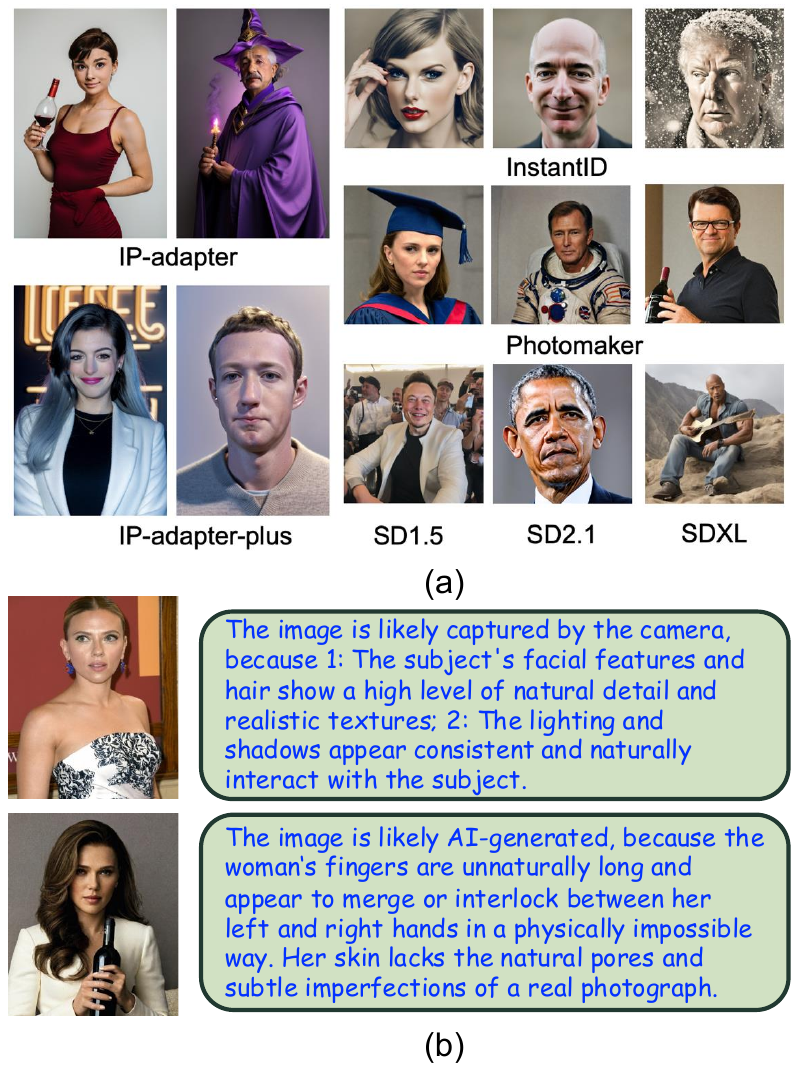} 
    \caption{(a) Examples of celebrity images generated by different diffusion personalized models. (b) Given the image, we use the Gemini-$2.0$ Pro to produce corresponding text annotations.}
    \label{fig:hufor_dataset}
\end{figure}
\begin{table}[t]
\centering
\label{tab:dataset_stats}
\begin{tabular}{c|ccc|c}
\toprule
\textbf{Dataset} & \textbf{FS} & \textbf{PM} & \textbf{Full body} & \textbf{Image $\#$} \\
\midrule
Uni-attack+ & \checkmark & \checkmark &  & $344,986$ \\
FF++ &  & \checkmark & \checkmark & $360,000$ \\
Diff-Cele & \checkmark &  & \checkmark & $317,231$ \\
\hline
HuFor & \checkmark & \checkmark & \checkmark & $1,022,217$ \\
\bottomrule
\end{tabular}
\caption{Statistics of the HuFor dataset. [Key: FS: fully-synthesized; PM: partially-manipulated].}
\label{tab:datasets_overview}
\end{table}

\subsection{Training and Inference}
\label{subsec:training_inference}
The training is a three-stage procedure that progressively integrates different components, ensuring stable optimization.

The first stage trains the $\mathcal{F}_{\text{ctx}}$ to generate textual rationales for its forgery decisions. 
The confidence token and $\mathcal{F}_{\text{face}}$ are deactivated at this stage. 
We employ Low-Rank Adaptation (LoRA) for efficient fine-tuning.
The model is trained on the HuFor dataset (Sec.~\ref{sec:dataset}), which comprises image-text pairs denoted as $\mathcal{D} = \{ (\mathbf{I}^{(i)}, \mathbf{Y}^{(i)} ) \}_{i=1}^{N}$, where $\mathbf{Y} = [y_1, y_2, ..., y_T]$ represents the token sequence of the target rationale. 
Examples are shown in Fig.~\ref{fig:hufor_dataset}\textcolor{red}{b}. 
The training objective minimizes the negative log-likelihood of next-token prediction:
\begin{equation}
    \mathcal{L} = -\mathbb{E}_{(\mathbf{I}, \mathbf{Y}) \sim \mathcal{D}} \left[ \sum_{t=1}^{T} \log P(y_t | \mathbf{I}, y_{<t}; \Theta_{\text{Lora}}) \right].
\end{equation}
\begin{table*}[t]
\centering
{\setlength{\tabcolsep}{3pt}\renewcommand{\arraystretch}{0.95}%
\scalebox{0.8}{
\begin{tabular}{@{}>{\centering\arraybackslash}p{2.5cm}
                |cccccccc|cccccccc|cccccccc|cccc@{}}
\toprule
\multirow{3}{*}{Method}
& \multicolumn{8}{c|}{FF++}
& \multicolumn{8}{c|}{UniAttack+}
& \multicolumn{8}{c|}{Diff-Cele}
& \multicolumn{4}{c}{Overall} \\
\cmidrule(lr){2-9}\cmidrule(lr){10-17}\cmidrule(lr){18-25}\cmidrule(lr){26-29}
&
\multicolumn{2}{c}{AUC} & \multicolumn{2}{c}{Acc} & \multicolumn{2}{c}{TPR95} & \multicolumn{2}{c|}{TPR99} &
\multicolumn{2}{c}{AUC} & \multicolumn{2}{c}{Acc} & \multicolumn{2}{c}{TPR95} & \multicolumn{2}{c|}{TPR99} &
\multicolumn{2}{c}{AUC} & \multicolumn{2}{c}{Acc} & \multicolumn{2}{c}{TPR95} & \multicolumn{2}{c|}{TPR99} &
AUC & Acc & TPR95 & TPR99 \\
&
(\%)$\uparrow$ & & (\%)$\uparrow$ & & (\%)$\uparrow$ & & (\%)$\uparrow$ & &
(\%)$\uparrow$ & & (\%)$\uparrow$ & & (\%)$\uparrow$ & & (\%)$\uparrow$ & &
(\%)$\uparrow$ & & (\%)$\uparrow$ & & (\%)$\uparrow$ & & (\%)$\uparrow$ & &
(\%)$\uparrow$ & (\%)$\uparrow$ & (\%)$\uparrow$ & (\%)$\uparrow$ \\
\midrule
\rowcolor{gray!20}
F3Net~\cite{qian2020thinking}
& 79.16 & & 76.60 & & 36.42 & & 10.23 & & 82.45 & & 80.95 & & 45.86 & & 11.79 & & 81.77 & & 75.27 & & 50.03 & & 15.55 & & 81.27 & 79.66 & 51.82 & 10.33 \\
  
SBI$^{\mathbf{*}}$~\cite{shiohara2022detecting}
& 82.15 & & 80.65 & & 38.11 & & 12.11 & & 77.09 & & 75.59 & & 40.42 & & 10.11 & & 72.12 & & 66.68 & & 54.03 & & 15.68 & & 76.60 & 76.01 & 43.07 & 11.52 \\
\rowcolor{gray!20}
RECCE\textbf{$^{*}$}~\cite{cao2022end}
& 83.30 & & 86.60 & & 41.11 & & 16.89 & & 75.60 & & 76.50 & & 31.88 & & 5.51 & & 69.55 & & 63.80 & & 56.02 & & 11.52 & & 75.61 & 72.69 & 41.09 & 8.45 \\

M2F2-Det~\cite{guo2025rethinking}
& \second{87.20} & & \second{85.70} & & \second{50.37} & & 21.79 & & 86.01 & & 84.51 & & 55.28 & & 12.54 & & 90.40 & & 88.90 & & 73.02 & & 34.11 & & 86.73 & 84.37 & 54.36 & 12.11 \\
\rowcolor{gray!20}
CNN-Det~\cite{wang2020cnn}
& 79.50 & & 76.60 & & 39.71 & & 17.77 & & 83.55 & & 80.02 & & 49.00 & & 9.87 & & 88.20 & & 86.70 & & 65.02 & & 27.83 & & 83.06 & 79.76 & 50.24 & 7.92 \\

UniFD$^{*}$~\cite{ojha2023towards}
& 70.20 & & 66.90 & & 41.07 & & 17.56 & & 83.17 & & 81.67 & & 51.77 & & 11.58 & & 94.85 & & 93.35 & & 75.92 & & 38.91 & & 82.18 & 79.43 & 53.60 & 10.90 \\
\rowcolor{gray!20}
NPR~\cite{tan2024rethinking}
& 82.14 & & 80.94 & & 49.09 & & \second{22.07} & & \second{86.09} & & \second{84.59} & & \second{63.18} & & \second{25.40} & & \best{\textbf{95.40}} & & \best{\textbf{93.90}} & & \best{\textbf{83.04}} & & \best{\textbf{53.89}} & & \second{87.75} & \second{87.98} & \second{64.99} & \second{24.15} \\
\midrule
HuForDet
& \best{\textbf{87.80}} & & \best{\textbf{86.30}} & & \best{\textbf{55.00}} & & \best{\textbf{28.04}} & &
  \best{\textbf{90.70}} & & \best{\textbf{89.20}} & & \best{\textbf{70.31}} & & \best{\textbf{35.81}} & &
  \second{95.10} & & \second{93.60} & & \second{80.07} & & \second{49.64} & &
  \best{\textbf{90.22}} & \best{\textbf{89.70}} & \best{\textbf{70.87}} & \best{\textbf{33.45}} \\
\bottomrule
\end{tabular}
}}
\caption{Detection performance on the HuFor dataset. \textcolor{red}{Best} and \textcolor{blue}{Second Best} are highlighted. $^\textbf{*}$ indicates that we apply released pre-trained weights. All metrics are reported as percentages.}
\label{tab:inter-dataset-protocol}
\end{table*}
The second stage focuses on training $\mathcal{F}_{\text{face}}$ using cropped facial regions, where $\mathcal{F}_{\text{face}}$ is optimized with binary cross-entropy loss:
\begin{equation}
    \mathcal{L}_{\text{face}} = -\mathbb{E}_{(\mathbf{I}_{\text{face}}, y_{\text{gt}})} \left[ y_{\text{gt}} \log y + (1 - y_{\text{gt}}) \log (1 - y) \right],
    \label{eq_ce}
\end{equation}
where $y = \mathcal{F}_{\text{face}}(\mathbf{I}_{\text{face}})$ is $\mathcal{F}_{\text{face}}$'s prediction and $y_{\text{gt}}$ is the ground-truth forgery label.

In the final stage, we freeze the pre-trained $\mathcal{F}_{\text{ctx}}$ and $\mathcal{F}_{\text{face}}$ branches, and then obtain their representations, \textit{i.e.}, $\mathbf{f}_{\text{ctx}}$ and $\mathbf{f}_{\text{face}}$.
We compute the confidence score $c$ based on Eq.~\ref{eq:confidence_score}.
The fusion model $\mathcal{G}$ takes the confidence-weighted concatenated features $[\mathbf{f}_{\text{ctx}}, c \cdot \mathbf{f}_{\text{face}}]$ as input and produces a predicted label $y_{\text{final}}$: $y_{\text{final}} = \mathcal{G}([\mathbf{f}_{\text{ctx}}, c \cdot \mathbf{f}_{\text{face}}])$.
The optimization is achieved using cross-entropy loss similar to Eq.~\ref{eq_ce}.

\noindent\textbf{Inference.}
During inference, given an input image $\mathbf{I}$, we first use the \texttt{dlib} library to crop the facial region $\mathbf{I}_{\text{face}} = \mathcal{C}_{\text{dlib}}(\mathbf{I})$. 
The full image $\mathbf{I}$ and the cropped face $\mathbf{I}_{\text{face}}$ are then fed into $\mathcal{F}_{\text{ctx}}$ and $\mathcal{F}_{\text{face}}$, respectively. 
Then, the final forgery probability is computed by the fusion network $\mathcal{G}$ as defined in Eq.~\ref{eq:fusion}.
\vspace{3mm}
\subsection{HuFor Dataset}
\label{sec:dataset}
We construct a human image forgery (HuFor) benchmark, a large-scale dataset that covers the full spectrum of manipulation techniques. HuFor is curated from three primary sources: 
(1) the widely-used FaceForensics++ (FF++) dataset~\cite{rossler2019faceforensics++v2}, which provides partially manipulated facial videos with different compression rates; 
(2) a diverse set of digital forgery images sourced from the UniAttackData+ benchmark~\cite{liu2025benchmarking}; 
and (3) as shown in Fig.~\ref{fig:hufor_dataset}, a novel corpus of fully-synthesized celebrity images generated by SoTA diffusion personalized models (Diff-Cele), including InstantID~\cite{wang2024instantid}, PhotoMaker~\cite{li2024photomaker}, and IP-Adapter~\cite{ye2023ip}. 
These diffusion personalized images bridge the critical gap in full-body forgeries. 
Specifically, we employed personalized diffusion models to generate images of over 30 distinct celebrity identities, each performing 15 different activities (\textit{e.g.}, running, playing an instrument, sitting at a laptop) across varied environments. 
This controlled generation process exposes consistent construction artifacts inherent to current generative models like implausible limb articulations.
In total, HuFor contains over 28 distinct forgery types with 1,022,217 images, as shown in Tab.~\ref{tab:datasets_overview}, encompassing both traditional partial manipulations (\textit{e.g.}, face-swapping, reenactment) and modern full syntheses from GANs and diffusion models.
The dataset is partitioned into training (30\%), validation (10\%), and testing (50\%) sets, containing 306,665; 102,222; and 511,109 images respectively.
More detailed dataset statistics are shown in the supplementary. 
\vspace{3mm}
\section{Experiment}
\label{sec:exp}

\subsection{Setup} 
\label{exp:setup}

\Paragraph{Dataset} We evaluate methods on three datasets:
\textbf{HuFor}, as described in Sec.~\ref{sec:dataset}, serves as our main benchmark for evaluating generalized human image forgery detection;
\textbf{FaceForensics++ (FF++)} is a standard benchmark for facial manipulation detection, containing $1,000$ original videos manipulated by four different forgery methods; 
\textbf{Celeb-DF} dataset~\cite{li2019celeb} features high-quality forgeries with fewer visible artifacts.

\Paragraph{Metrics} We report performance using four metrics: the Area Under the Curve (AUC), Accuracy, and the True Positive Rates at 5\% and 1\% False Positive Rates (TPR95 and TPR99). Specifically, accuracy is computed using the optimal threshold that maximizes classification performance, while TPR95 and TPR99 evaluate detection capabilities under increasingly stringent false alarm tolerances, reflecting practical deployment reliability where minimizing false positives is critical.

\Paragraph{Implmentation Details} 
We use a DenseNet-121~\cite{tan2019efficientnet} as a baseline of $\mathcal{F}_{face}$, with the proposed MoE layer integrated between the $3$rd and $6$th convolutional blocks; the RGB domain expert is built upon on DenseNet blocks.
$\mathcal{F}_{ctx}$ leverages a CLIP-ViT/336px vision encoder and a Vicuna-7B large language model (LLM), for which we expand the vocabulary with a single special token \texttt{<s>} as a dedicated confidence token. 
For input processing, the \texttt{dlib} package detects facial regions from all images, retaining a maximum of five largest faces per image. 
Complete implementation details are in the supplementary.

\subsection{Performance on HuFor Dataset}
\label{ex:detection}
Tab.~\ref{tab:inter-dataset-protocol} shows that our HuForDet achieves SoTA performance on the HuFor dataset, with the highest overall AUC of 90.22\% --- a significant improvement of +2.47\% over NPR (87.75\%) and +3.49\% over M2F2-Det (86.73\%).
Also, HuForDet has substantial improvements on overall TPR95 of 70.87\% (+5.88\% over NPR) and TPR99 of 33.45\% (+9.30\% over NPR), highlighting its superior performance under low false-positive constraints.

Specifically, on the FF++ subset, which primarily contains partial facial manipulations, HuForDet achieves competitive performance (87.80\% AUC) with methods like M2F2-Det (87.20\% AUC) that utilize sophisticated forgery masks and additional forgery detection components. 
We also provide additional FF++ detection results and analysis in Sec.~\ref{sec:deefake}.
More notably, HuForDet demonstrates remarkable superiority on the UniAttack+ subset, achieving the highest AUC of 90.70\%. 
This represents substantial improvements of +4.61\% over NPR (86.09\%) and +2.69\% over M2F2-Det (86.01\%).
Also, our method's advantage is more clear when measured by TPR99, achieving +10.41\% over NPR (25.40\%).
The UniAttack+ dataset combines both partially-manipulated and fully-synthesized forgeries.
As a result, UniFD's vision-language approach lacks the fine-grained local analysis needed for subtle facial manipulations, while traditional face detectors like SBI (77.09\% AUC) cannot handle full-body synthetic samples. 
In contrast, HuForDet is a holistic method that effectively handles this via the face and contextual forgery detection branches, which identify face-region forgery and semantic inconsistencies in synthesized whole-body images, respectively.
In addition, our confidence-guided fusion further optimizes this collaboration, achieving effective detection across all attack types, which will be detailed in Sec.~\ref{sec:visualization}.
However, on the Diff-Cele subset containing fully-synthesized images, HuForDet (95.10\%) is slightly outperformed by NPR (95.40\%). 
This is expected, as NPR specializes in identifying local interdependencies among image pixels induced by upsampling operators in generative models—a characteristic strongly evident in synthetic images from GANs or diffusion models. 
Nevertheless, NPR's specialized design becomes a limitation when dealing with partial manipulations, where only a portion of pixels is forged, and the local interdependence signal becomes insufficient for reliable detection. 
This explains NPR's comparatively weaker performance on FF++ (82.14\% AUC) and UniAttack+ (86.09\% AUC).
In contrast, HuForDet's holistic approach effectively handles both partial manipulations and fully-synthesized images, demonstrating robust performance across diverse forgery types.

\subsection{Ablation Study} 
\begin{table}[t]
\centering
\scalebox{0.9}{
\begin{tabular}{lc|cccc}
\hline
&\textbf{Model Variant} & \textbf{FF++} & \textbf{Uni.} & \textbf{DC.} & \textbf{Overall} \\
\hline
\rowcolor{gray!20}1& Baseline & 83.51 & 80.14 & 83.85 & 82.50 \\
\hline
2& + $E_{rgb}$ & 83.02 & 82.25 & 84.22 & 82.93 \\
\hline
\rowcolor{gray!20}3& + $E_{freq}$ & 89.10 & 86.63 & 85.00 & 86.88 \\
\hline
4& + $E_{rgb}$ + $E_{freq}$ & 90.95 & 87.20 & 88.11 & 90.42 \\
\hline
\rowcolor{gray!20}5& + $\mathcal{F}_{ctx}$ (concat) & 69.80 & 78.95 & 79.50 & 75.75 \\
\hline
6& + $\mathcal{F}_{ctx}$ (fuse) & \textbf{91.01} & \textbf{88.60} & \textbf{91.05} & \textbf{91.45} \\
\hline
\end{tabular}}
\caption{Ablations on the HuFor validation set. Performance is measured by AUC(\%). [Key:  Uni.: UniAttack+; DC: Diff-Cele; $E_{rgb}$: {$E_{1}$ and $E_{2}$}; $E_{freq}$: {$E_{3}$ and $E_{4}$}].}
\label{tab:ablation_detailed}
\end{table}

Tab.~\ref{tab:ablation_detailed} begins with a baseline (\textit{i.e.}, DenseNet-121 in \textbf{Row 1}), which achieves 82.50\% overall AUC.
The introduction of RGB domain experts (\textbf{Row 2}) shows modest but targeted improvements, particularly on the UniAttack+ subset (82.25\%), indicating its effectiveness in detecting forgeries in fully-synthesized images.
In contrast, frequency domain experts (\textbf{Row 3}) demonstrate substantially stronger performance, elevating overall AUC to 86.88\% and excelling on FF++ (89.10\%) by effectively capturing high-frequency blending artifacts. 
\textbf{Row 4} shows that the combined MoE framework achieves 90.42\% overall AUC, substantially outperforming either standalone expert and demonstrating their complementarity.
Additionally, the integration methodology for the contextualized forgery detection branch, \textit{i.e.}, $\mathcal{F}_{ctx}$, proves critically important. 
Naive feature concatenation (\textbf{Row 5}) causes severe performance degradation to 75.75\% overall AUC, particularly on FF++ (69.80\%). 
This failure stems from the fact that, when semantic reasoning lacks visual evidence, high-dimensional MLLM embeddings can be misleading. 
Our proposed confidence-aware dynamic fusion (\textbf{Row 6}) successfully resolves this conflict by learning to weight branch contributions based on input forgery types, achieving the best overall performance of 91.45\% AUC --- 15.70\% improvement over naive concatenation, which further demonstrates the necessity of a dynamic fusion.

\begin{figure}[t]
    \centering
    \includegraphics[width=1\linewidth]{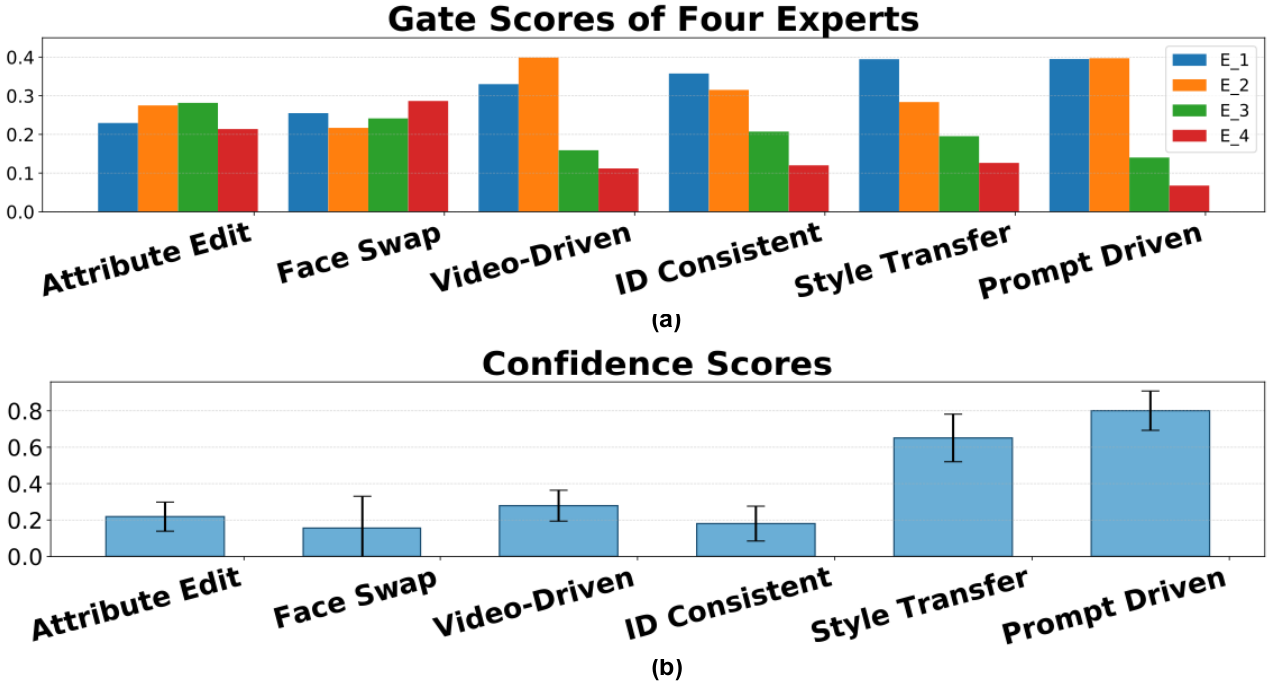} 
    \caption{
    Analysis of (a) gate scores and (b) confidence scores across six forgery categories, which are defined in~\cite{liu2025benchmarking} for digital generation and manipulation.
    }
    \label{fig:prompt_design}
\end{figure}
\begin{table}[t]
\centering
\scalebox{0.9}{
\begin{tabular}{l|c|cccc}
\hline
& $\sigma$ & \textbf{FF++} & \textbf{Uni.} & \textbf{DC.} & \textbf{Overall} \\
\hline
\rowcolor{gray!20}1& Baseline & 83.51 & 80.14 & 83.85 & 82.50 \\
\hline
2& $\{1,4,7\}$   & 87.20 & 83.10 & 84.20 & 84.83 \\
\rowcolor{gray!20}3& $\{9,12,15\}$ & 85.10 & 82.20 & 84.90 & 84.07 \\
4& $\{1,4,7, 9,12,15\}$ & 88.00 & 85.10 & 84.60 & 85.90 \\
\hline
5& + Ada-LoG & \textbf{89.10} & \textbf{86.63} & \textbf{85.00} & \textbf{86.88} \\
\hline
\end{tabular}}
\caption{The adaLoG block analysis on the HuFor validation set. Performance is measured by AUC(\%). [Key: Uni.: UniAttack+; DC: Diff-Cele; $\sigma$ controls the blurring scale in LoG operators.]\vspace{-3mm}}
\label{tab:adalog}
\end{table}

\begin{figure*}[t]
\centering
\begin{subfigure}[b]{0.21\textwidth}
    \centering
    \includegraphics[width=1\linewidth]{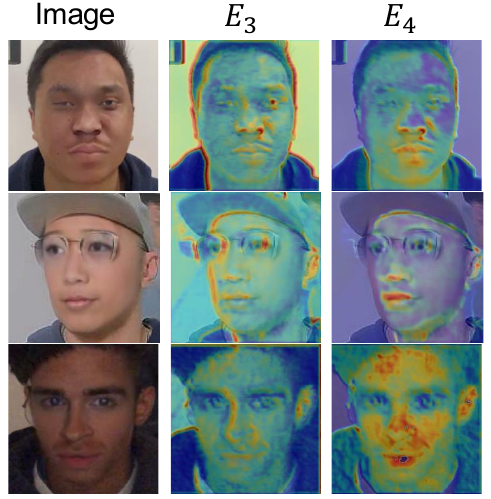}
    \subcaption{}
    \label{fig:gallery}
\end{subfigure}
\hfill
\hspace{-3mm}
\begin{subfigure}[b]{0.33\textwidth}
    \centering
    \includegraphics[width=0.95\linewidth]{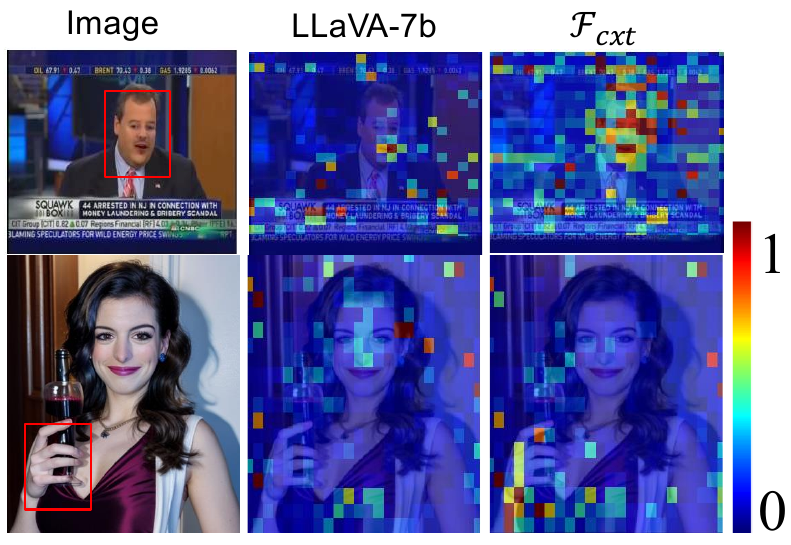}
    \subcaption{}
    \label{fig:attn_viz}
\end{subfigure}
\hfill
\hspace{-3mm}
\begin{subfigure}[b]{0.42\textwidth}
    \centering
    \scalebox{0.87}{
    \begin{tabular}{c|c|c|c}
        \hline
        \multirow{2}{*}{Methods} & \multicolumn{1}{c|}{FF++ (c23)} & \multicolumn{1}{c|}{FF++ (c40)} & \multicolumn{1}{c}{Celeb-DF} \\
        \cline{2-4}
        & \multicolumn{3}{c}{\textit{Metric:} Acc (\% $\uparrow$) / AUC (\% $\uparrow$)}\\
        \hline
        Add-Net~\cite{zi2020wilddeepfake} & $96.78$ / $97.74$ & $87.50$ / $91.01$ & $96.93$ / $99.55$ \\
        \rowcolor{gray!30}F3-Net~\cite{qian2020thinking} & $97.52$ / $98.10$ & $90.43$ / $93.30$ & $95.95$ / $98.93$ \\
        \rowcolor{gray!30}RECCE~\cite{cao2022end} & $97.06$ / $99.32$ & $91.03$ / {\color{blue}95.02} & {\color{blue}98.59} / {\color{blue}99.94} \\
        TALL~\cite{xu2023tall} & {\color{blue}98.65} / {\color{red}\textbf{99.87}} & $92.82$ / $94.57$ & $97.57$ / $98.55$ \\
        \rowcolor{gray!30}LAA~\cite{nguyen2024laa} & $97.06$ / $99.32$ & $91.03$ / -- & -- \\
        M2F2~\cite{m2f2_det_guo} & $98.79$ / $99.34$ & {\color{red}\textbf{93.83}} / {\color{red}\textbf{96.58}} & $98.98$ / $99.02$ \\
        \hline
        \rowcolor{gray!30}HuForDet & {\color{red}\textbf{99.11}} / {\color{blue}99.44} & {\color{blue}92.99} / {\color{blue}95.21} & {\color{red}\textbf{99.01}} / {\color{red}99.96} \\
        \hline
    \end{tabular}}
    \subcaption{}
    \label{tab:intra-dataset-protocol}
\end{subfigure}
\hspace{+3mm}
\vspace{-3mm}
\caption{(a) Visualizations on face regions and feature maps obtained via adaLoG blocks from $E_3$ and $E_4$, respectively. (b) The $\mathcal{F}_{\text{ctx}}$ focuses on forged regions such as facial manipulations in the first example and anomalous finger artifacts in the second. (c) Face swap detection performance.\vspace{-3mm}}
\label{fig:one_row_table_two_figs_subfig}
\end{figure*}

\subsection{Analysis and Visualizations} 
\label{sec:visualization}
\paragraph{Expert Gate Scores} Fig.~\ref{fig:prompt_design}\textcolor{red}{a} shows that two RGB domain experts ($E_1$ and $E_2$) receive higher gate scores than frequency-domain experts ($E_3$ and $E_4$), identifying them as important contributors to HuForDet’s detection capability.
However, frequency-domain experts show a marked increase in their relative importance for partial manipulations such as Face Swap and Attribute Edit, where their individual contributions rise to approximately equal levels as RGB domain experts.
This indicates that the model dynamically leverages its full suite of specialized experts, relying on the foundational detection of RGB-domain experts while recruiting additional capacity from frequency-domain experts to handle the complex, localized artifacts unique to partial manipulations.

\Paragraph{Adaptive LoG Block}~Tab.~\ref{tab:adalog} provides analysis on the adaLoG block, which serves as our frequency-domain experts.
Specifically, the fixed small-scale LoG (\textbf{Row 2}) yields better results (84.83\% AUC) than coarse-scale LoG (\textbf{Row 3}) (84.07\% AUC), suggesting that fine-grained artifacts are more discriminative for forgery detection. 
Then, \textbf{Row 4} shows that the combination of different scales further improves performance to 85.90\% AUC.
Importantly, our adaLoG achieves the highest overall performance of 86.88\% AUC, outperforming all fixed-scale configurations --- particularly a strong gain on FF++ (89.10\% AUC). 
This performance enhancement validates our hypothesis that spatially adaptive scale selection is crucial in detection, and our adaLoG block effectively learns optimal scale representations. 

\Paragraph{Contextualized Detection Branch Confidence} Fig.~\ref{fig:prompt_design}\textcolor{red}{b} shows that confidence scores from the $\mathcal{F}_{\text{ctx}}$ exhibit distinct distributions across forgery types.
Specifically, prompt-driven, a fully-synthetized forgery type, exhibits consistently high confidence scores with a mean of 0.80, indicating $\mathcal{F}_{\text{ctx}}$ has a reliable detection in this forgery category.
In contrast, Face Swap forgeries show lower confidence (mean: 0.18) with large variance, reflecting the challenge of identifying forgeries when only local regions are manipulated within other authentic contexts. 
These statistics show that our confidence mechanism weights $\mathcal{F}_{\text{ctx}}$'s contributions based on different forgery types.

\Paragraph{Visualizations} 
First two rows of Fig.~\ref{fig:gallery} demonstrate that $E_{3}$ strongly activates on the manipulated eye and eyeglass regions, which are fine-scale artifacts our adaLoG learns to capture with a smaller $\sigma$.
In contrast, when obvious artifacts are absent (third row), $E_{4}$ exhibits broader responses than $E_{3}$, suggesting the forgery manifests primarily in facial textures.
Also, Fig.~\ref{fig:attn_viz} shows the learned behavior of $\mathcal{F}_{ctx}$ through cross-modality attention visualization. 
By aggregating attention maps across LLM's transformer layers, we observe that our $\mathcal{F}_{ctx}$ effectively focuses on forgery regions, such as facial areas in the face-swap example and anomalous finger artifacts in the second image. 
This contrasts with the original LLaVA-7b model, which fails to produce meaningful forgery attention maps due to its lack of specialized detection training. 

\subsection{Face Forensic Dataset Performance} 
\label{sec:deefake}
Tab.~\ref{tab:intra-dataset-protocol} demonstrates HuForDet's competitive capability in traditional face-swap detection. 
On the FF++ c23 dataset, our method achieves SoTA accuracy of 99.11\% and AUC of 99.44\%, outperforming a strong frequency-based method like F3-Net. 
On the more challenging FF++ c40 dataset, where artifacts are less visible, HuForDet maintains robust performance with 92.99\% accuracy and 95.21\% AUC, slightly worse than M2F2-Det. 
Furthermore, HuForDet's strong performance on the challenging Celeb-DF dataset (99.01\% accuracy) again confirms its effectiveness in identifying facial region foregeries.
\section{Conclusion}
We introduce HuForDet, a holistic detection method for human image forgery.
By combining a face forgery detection branch with heterogeneous experts --- including a novel adaptive LoG for multi-scale frequency analysis --- with a contextualized forgery detection branch that leverages MLLM reasoning and confidence-aware fusion, our HuForDet captures both localized facial artifacts and global semantic anomalies.
Our HuForDet achieves SoTA performance by effectively generalizing across both partial manipulations and full synthesized human image forgeries.
More importantly, this work provides a foundation for defending against evolving AI-generated human image forgeries, with future work aimed at improving efficiency.
\clearpage
{
    \small
    \bibliographystyle{ieeenat_fullname}
    \bibliography{main}
}
\end{document}